\documentclass[12pt]{article}

\usepackage{amsmath}
\usepackage{tikz-cd}
\usetikzlibrary{shapes.geometric}
\usepackage{amssymb}
\usepackage{graphicx}
\usepackage{hyperref}
\usepackage{geometry}
\usepackage{mathtools}
\usepackage{titlesec} 
\usepackage{titling}
\usepackage[backend=bibtex,style=numeric,sorting=none]{biblatex}
\usepackage{etoolbox}   
\usepackage{booktabs}   
\usepackage{subcaption} 
\usepackage{multirow}   
\usepackage{xcolor}     
\graphicspath{{./experiment_results/}}

\titleformat{\section}      {\normalfont\Large\bfseries\centering}{\thesection}{1em}{}
\titleformat{\subsection}   {\normalfont\large\bfseries\centering}{\thesubsection}{1em}{}
\titleformat{\subsubsection}{\normalfont\normalsize\bfseries\centering}{\thesubsubsection}{1em}{}
\titleformat{\paragraph}[block]{\normalfont\normalsize\bfseries\centering}{}{0em}{}
\titlespacing*{\paragraph}{0pt}{1.5ex plus 0.5ex minus .2ex}{0.5ex}

\titleclass{\subsubsubsection}{straight}[\subsubsection]
\newcounter{subsubsubsection}[subsubsection]
\renewcommand\thesubsubsubsection{\thesubsubsection.\arabic{subsubsubsection}}
\titleformat{\subsubsubsection}
  {\normalfont\normalsize\bfseries\centering}{\thesubsubsubsection}{1em}{}
\titlespacing*{\subsubsubsection}{0pt}{1.5ex plus 0.5ex minus .2ex}{1em}

\pretitle{%
  \begin{center}
    \vspace*{-0.42cm}
    \rule{\linewidth}{3pt}\\[\smallskipamount]
    \bfseries\LARGE
    \vspace{0.22cm}
}
\posttitle{%
    \\[\smallskipamount]
    \rule{\linewidth}{2pt}
  \end{center}
  \vspace{0.12cm}
}

\preauthor{\begin{center}\large}
\postauthor{\end{center}}

\predate{\begin{center}\large}
\postdate{\end{center}}

\makeatletter
\g@addto@macro\maketitle{\thispagestyle{empty}}
\makeatother

\title{General Proximal Flow Networks}

\author{%
  \begin{minipage}[t]{0.46\textwidth}\centering
    Alexander Strunk\thanks{Corresponding author: \href{mailto:astrunk.research@evercot.ai}{astrunk.research@evercot.ai}}\\
    Evercot AI\\
  \end{minipage}%
  \hspace{1em}
  \begin{minipage}[t]{0.46\textwidth}\centering
    Roland Assam\\
    Evercot AI\\
  \end{minipage}%
\vspace{0.42cm}
}

\date{19 January 2026}

\begin{document}

\maketitle  

\vspace{-0.80cm} 
\begin{abstract}
\noindent
This paper introduces \emph{General Proximal Flow Networks} (GPFNs), a generalization of Bayesian Flow Networks~\cite{graves2023bfn} that broadens the class of admissible belief-update operators.
In Bayesian Flow Networks, each update step is a Bayesian posterior update, which is equivalent to a proximal step with respect to the Kullback--Leibler divergence.
GPFNs replace this fixed choice with an arbitrary divergence or distance function, such as the Wasserstein distance, yielding a unified proximal-operator framework for iterative generative modeling.
The corresponding training and sampling procedures are derived, establishing a formal link to proximal optimization and recovering the standard BFN update as a special case.
Empirical evaluations confirm that adapting the divergence to the underlying data geometry yields measurable improvements in generation quality, highlighting the practical benefits of this broader framework.
\end{abstract}

\newpage
\setcounter{page}{1}

\section{Introduction}
Deep generative modeling has advanced rapidly through frameworks such as Diffusion Models~\cite{nicholdh2021improved} and Flow Matching~\cite{lipman2022flow}, which rely on \emph{iterative refinement} to transform simple noise into complex data.
Bayesian Flow Networks (BFNs)~\cite{graves2023bfn} approach this process differently: instead of transforming samples directly, they evolve a \emph{belief distribution} over the data space via sequential Bayesian posterior updates.
At each step, a neural network predicts the underlying data distribution to guide the generation process, while the belief updates themselves remain strictly independent of these predictions during training.
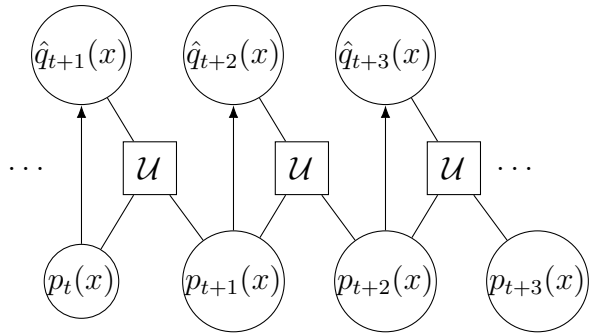
\begin{figure}[htpb!]
\begin{center}
    \begin{tikzpicture}[
  >=Latex,
  circ/.style={draw, circle, minimum size=8mm, inner sep=0pt},
  sq/.style={draw, rectangle, minimum size=7mm, inner sep=0pt},
  line/.style={draw},
  arr/.style={draw, ->}
]

\def\xsep{3.0cm}
\def\ysep{2.0cm}

\def\ydotsR{-3*\ysep+0.3cm}         
\def\ydotsL{\ysep-1.27cm}  
\def\xdotsR{\xsep/2}         

\begin{scope}[rotate=90]

\node[circ] (L1) at (0, 0) {$p_t(x)$};
\node[circ] (R1) at (\xsep, 0) {$\hat{q}_{t+1}(x)$};
\node[sq]   (S1) at (\xsep/2, -0.9cm) {$\mathcal{U}$};

\draw[arr]  (L1) -- (R1);
\draw[line] (L1) -- (S1);
\draw[line] (R1) -- (S1);

\node[circ] (L2) at (0, -\ysep) {$p_{t+1}(x)$};
\node[circ] (R2) at (\xsep, -\ysep) {$\hat{q}_{t+2}(x)$};
\node[sq]   (S2) at (\xsep/2, -\ysep-0.9cm) {$\mathcal{U}$};

\draw[line] (S1) -- (L2);
\draw[arr]  (L2) -- (R2);
\draw[line] (L2) -- (S2);
\draw[line] (R2) -- (S2);

\node[circ] (L3) at (0, -2*\ysep) {$p_{t+2}(x)$};
\node[circ] (R3) at (\xsep, -2*\ysep) {$\hat{q}_{t+3}(x)$};
\node[sq]   (S3) at (\xsep/2, -2*\ysep-0.9cm) {$\mathcal{U}$};

\draw[line] (S2) -- (L3);
\draw[arr]  (L3) -- (R3);
\draw[line] (L3) -- (S3);
\draw[line] (R3) -- (S3);

\node[circ] (L4) at (0, -3*\ysep) {$p_{t+3}(x)$};
\draw[line] (S3) -- (L4);

\node at (\xdotsR, \ydotsR) {$\hdots$};
\node at (\xdotsR, \ydotsL) {$\hdots$};

\end{scope}

\end{tikzpicture}
\caption{The GPFN generation process.}
\label{fig:generation_process}
\end{center}
\end{figure}

\noindent Although mathematically elegant, the BFN posterior update is functionally equivalent to a proximal point step that is restricted to the Kullback--Leibler (KL) divergence.
This implicit geometric constraint can prove suboptimal for structured domains like images, where alternative distance measures, such as the Wasserstein metric, more naturally capture the underlying data geometry.

\noindent To address this limitation, \emph{General Proximal Flow Networks} (GPFNs) are introduced. This framework generalizes the belief update by replacing the rigid constraint of the KL divergence with an arbitrary distance function $D$.
The resulting update operator $\mathcal{U}$ solves a regularized optimization problem, effectively balancing fidelity to the predicted data distribution against proximity to the current belief state.
This provides a flexible framework that not only subsumes standard BFNs but also enables custom update rules tailored to specific data geometries.

\noindent The generation process of a GPFN can be described as follows: At each time step $t$, the current belief $p_t(x)$ is fed into a neural network that predicts the subsequent target distribution $\hat{q}_{t+1}(x)$.
The proposed update operator $\mathcal{U}$ then combines this prediction with $p_t(x)$ to yield the refined belief $p_{t+1}(x)$, forming the starting point for the subsequent step.
This recurrent architecture allows the model to progressively sharpen its belief regarding the data over $T$ iterations.
The corresponding training procedure, wherein the belief is updated using the \emph{true} data distribution $q(x)$ rather than the network's prediction.
\begin{figure}[htpb]
\begin{center}
    \begin{tikzpicture}[
  >=Latex,
  circ/.style={draw, circle, minimum size=8mm, inner sep=0pt},
  bigcirc/.style={draw, circle, minimum size=13mm, inner sep=0pt},
  sq/.style={draw, rectangle, minimum size=7mm, inner sep=0pt},
  diamond/.style={draw, rectangle, minimum size=9mm, inner sep=0pt, rotate=45},
  gate/.style={draw, fill=white, regular polygon, regular polygon sides=3, minimum size=7mm, inner sep=0pt},
  thin/.style={line width=0.8pt},
  thick/.style={line width=1.2pt},
  arr/.style={->, line width=0.8pt}
]

\begin{scope}[rotate=90, xscale=1.5, yscale=2.0]

\node[circ]    (A) at (0.0,  2.2) {$p_t(x)$};   
\node[circ]    (B) at (3.2,  2.2) {$\hat{q}_{t+1}(x)$};   
\node[bigcirc] (C) at (6.4,  0.6) {$q$};   

\node[diamond] (D) at (2.4,  1.1) {\rotatebox{-45}{$\mathcal{U}$}};   

\node[circ]    (E) at (0.0,  0.0) {$p_{t+1}(x)$};   
\node[circ]    (F) at (2.8,  0.1) {$\hat{q}_{t+2}(x)$};  

\node[diamond]      (S) at (3.3, -0.75) {\rotatebox{-45}{$\mathcal{U}$}};  
\node[circ]    (H) at (0.0, -2.2) {$p_{t+2}(x)$};   

\coordinate (T1) at (4.6, 1.45);
\coordinate (T2) at (4.0, 0.27);

\draw[arr]  (A) -- (B);                 
\draw[thick](A) -- (D);                 
\draw[arr](D) -- (E);                 
\draw[thick] (D) -- (C);                 

\draw[thin] (B) -- (T1);                
\draw[thin] (T1) -- (C);                

\draw[arr] (E) -- (F);                 
\draw[arr]  (F) -- (C);                 
\draw[thin] (F) -- (T2);                

\draw[thin] (E) -- (S);                 
\draw[thick](S) -- (C);                 
\draw[thick](H) -- (S);                 

\node[gate, rotate=00] at (T1) {$\mathcal{D}$}; 
\node[gate, rotate=00] at (T2) {$\mathcal{D}$}; 

\node at (3.2, 3.15) {$\hdots$};
\node at (3.2,-1.85) {$\hdots$};

\end{scope}

\end{tikzpicture}
\caption{The GPFN training procedure.}
\label{fig:training_procedure}
\end{center}
\end{figure}
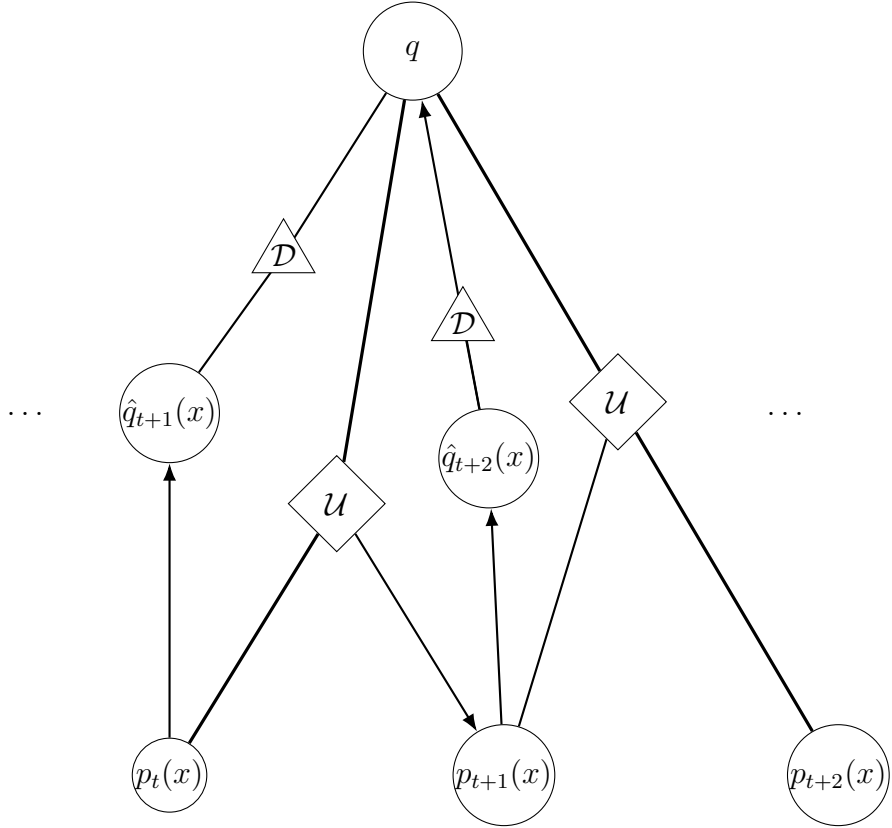
During training, the loss $\mathcal{D}$ measures the discrepancy between the neural network's output $\hat{q}_{t+1}$ and the true target signal $q_{t+1}$ at each step.
Critically, the \emph{true} target signal $q_{t+1}$---rather than the prediction $\hat{q}_{t+1}$---is passed directly into the update operator $\mathcal{U}$ to advance the belief state.
Consequently, the neural network predictions are utilized solely to define the training objective; they do not feed back into the belief trajectory during the optimization process.
This distinct separation between the learning signal and the underlying belief dynamics is inherited from the original BFN framework and is rigorously preserved within GPFNs.

\noindent The main contributions of this paper are structured as follows:
\begin{itemize}
    \item The General Proximal Flow Network framework is established, replacing the strictly KL-divergence-based proximal step found in BFNs with an arbitrary divergence or distance function. In parallel, the corresponding training and sampling procedures are rigorously derived (Section~\ref{sec:math}).
    \item A formal mathematical connection between GPFNs and proximal-point methods from convex optimization is formulated, demonstrating that baseline BFNs are recovered natively as a special case when employing the KL divergence.
    \item A Gaussian GPFN utilizing a Wasserstein-based update is evaluated against a standard BFN baseline on the MNIST dataset~\cite{lecun1998mnist}. This empirical analysis demonstrates improved generation quality, as measured by Fr\'echet Inception Distance (FID)~\cite{heusel2017fid} and related performance metrics (Section~\ref{sec:experiments}).
\end{itemize}

\newpage
\section{Mathematical Background}\label{sec:math}

A General Proximal Flow Network (GPFN) is a generative model that refines a belief distribution over the data space $\mathcal{X}$ through a sequence of $T$ proximal update steps.
Each step is governed by a divergence or distance function $D$ that may be chosen freely, with the Bayesian posterior update of standard BFNs recovered as the special case $D = \mathrm{KL}$.
The framework is specified by four components.

\paragraph{Belief distribution}
At each time step $t \in \{0, \ldots, T\}$, the model maintains a belief distribution
\[
  p_t \in \mathcal{P}(\mathcal{X}),
\]
where $\mathcal{P}(\mathcal{X})$ denotes the set of probability distributions over $\mathcal{X}$.
The process is initialized at $p_0$, a simple prior (e.g.\ a standard Gaussian), and terminates at $p_T$, which approximates a Dirac delta at a generated data point.

\paragraph{Target signal}
At each step $t$, a target signal $q_{t+1}$ is provided to guide the belief update.
In the general case, $q_{t+1}$ may depend on the time step $t$ and the true data $x_0$.
For example, in standard BFNs, $q_{t+1}$ corresponds to the likelihood of a noisy observation $y_{t+1}$ drawn from a time-dependent distribution centered at $x_0$.
In other instantiations, such as the $W_2$-GPFN, the target can simply be the fixed clean data $q_{t+1} = \delta_{x_0}$ for all $t$.
During training, this target signal $q_{t+1}$ is used to guide the belief updates and provide supervision.

\paragraph{Neural network predictor}
A neural network $F_\theta$ maps the current belief $p_t$ to a predicted target distribution:
\[
  \hat{q}_{t+1} = F_\theta(p_t) \in \mathcal{P}(\mathcal{X}).
\]
In practice, $F_\theta$ is parameterized as a denoising network that predicts the clean data $\hat{x}_0$ from the current belief, implicitly defining $\hat{q}_{t+1} = \delta_{\hat{x}_0}$.

\paragraph{Proximal update operator}
The belief is advanced from $p_t$ to $p_{t+1}$ by solving a regularized optimization problem that balances two competing objectives: \emph{fidelity} to the target signal $q_{t+1}$ (which is the true signal during training and the predicted signal $\hat{q}_{t+1}$ during sampling) and \emph{proximity} to the current belief $p_t$.
Formally,
\begin{equation}\label{eq:proximal-update}
  p_{t+1} = \mathcal{U}_t(p_t,\, q_{t+1})
  \;:=\; \arg\min_{p \,\in\, \mathcal{P}(\mathcal{X})} \Big[\, \mathcal{F}_t(p,\, q_{t+1}) + \frac{1}{\eta_t}\, D(p,\, p_t) \,\Big],
\end{equation}
where $\mathcal{F}_t$ is a fidelity functional measuring the discrepancy to the target, $D(\cdot,\cdot)$ is the chosen proximal divergence, and $\eta_t > 0$ is a step-size parameter controlling the trade-off between the two terms.
This is precisely the structure of a \emph{proximal-point step} from convex optimization \cite{parikh2014proximal}, applied here to the space of probability measures.
This formulation elegantly unifies different generative paradigms.
When $D = W_2^2$ (the squared $2$-Wasserstein distance) and $\mathcal{F}_t(p, q_{t+1}) = W_2^2(p, \delta_{x_0})$, equation~\eqref{eq:proximal-update} computes the Wasserstein barycenter between the current belief $p_t$ and the clean data. The solution corresponds exactly to McCann's displacement interpolation, moving $p_t$ along the optimal transport geodesic towards $\delta_{x_0}$ with an effective step size $\tau_t = \eta_t / (1 + \eta_t)$.
Conversely, when $D(p, p_t) = \mathrm{KL}(p \,\|\, p_t)$ and $\mathcal{F}_t$ is the expected negative log-likelihood $-\mathbb{E}_{x \sim p}[\ln L(x \mid y_{t+1})]$ of a noisy observation $y_{t+1}$, setting $\eta_t = 1$ yields the closed-form minimizer $p_{t+1}(x) \propto p_t(x) L(x \mid y_{t+1})$. This exactly recovers the Bayesian posterior update of standard BFNs.
By treating the divergence $D$ as a flexible design choice, the update rule can be explicitly adapted to the underlying geometry of the data space---allowing a shift from the pointwise, information-theoretic topology of the KL divergence to the spatial, mass-moving geometry of optimal transport.

\paragraph{Training}
During training, the true data $x_0$ is known, and the target signals $q_{t+1}$ are generated according to the chosen forward process.
The belief trajectory $p_0, p_1, \ldots, p_T$ is generated using these true targets via the update operator~\eqref{eq:proximal-update}:
\[
  p_{t+1} = \mathcal{U}_t\!\bigl(p_t,\; q_{t+1}\bigr).
\]
The parameters $\theta$ of the predictor $F_\theta$ are optimized to minimize the expected discrepancy between the predicted target $\hat{q}_{t+1}$ and the true target $q_{t+1}$, summed over all time steps:
\begin{equation}\label{eq:training-loss}
  \mathcal{L}(\theta) = \sum_{t=0}^{T-1} \mathbb{E}\bigl[\mathcal{D}\!\bigl(\hat{q}_{t+1},\; q_{t+1}\bigr)\bigr],
\end{equation}
where $\mathcal{D}$ is a suitable loss functional.
A key structural property of the framework is that the belief trajectory is generated by the \emph{true} targets $q_{t+1}$, not by the network's predictions $\hat{q}_{t+1}$.
The predictions enter only through the loss~\eqref{eq:training-loss}; they do not feed back into the belief updates.
This separation ensures that the belief trajectory is stable and independent of the current quality of $F_\theta$.

\paragraph{Sampling}
At generation time, the true targets $q_{t+1}$ are unavailable.
The network's predictions are therefore used in their place: starting from the prior $p_0$, the belief is updated iteratively as
\[
  p_{t+1} = \mathcal{U}_t\!\bigl(p_t,\; \hat{q}_{t+1}\bigr), \qquad t = 0, 1, \ldots, T-1,
\]
with $\hat{q}_{t+1} = F_\theta(p_t)$.
After $T$ steps, samples are drawn from the final belief $p_T$.
The quality of the generated samples depends on how accurately the trained predictor $F_\theta$ approximates the true targets, as well as on the suitability of the chosen proximal divergence $D$ for the geometry of the data.

\section{Experiments}\label{sec:experiments}

A Gaussian instantiation of the GPFN framework is compared against a standard Bayesian Flow Network (BFN)~\cite{graves2023bfn} on MNIST~\cite{lecun1998mnist}.
Both models share an identical U-Net backbone (${\approx}4\text{M}$ parameters) and training budget; the only difference is the generative framework and its update operator.
Full implementation details are given in Appendix~\ref{app:impl}.

\subsection{Setup}

\paragraph{GPFN}
The belief is parameterized as an axis-aligned Gaussian $p_t = \mathcal{N}(m_t, v_t I)$.
The proximal update~\eqref{eq:proximal-update} with $D = W_2^2$ yields a closed-form update for both the mean and the variance; the network $F_\theta$ is trained to predict the clean image $\hat{x}_0$ with pixel-level MSE.

\paragraph{BFN baseline}
The BFN baseline architecture and training objective follow Graves et al.~\cite{graves2023bfn}.
Two sampling schemes are evaluated: the original stochastic sampler~\cite{graves2023bfn} (BFN-stoch) and a custom deterministic $W_2$-proximal sampler (BFN-det) designed to allow direct comparison with GPFN-det.

\paragraph{Evaluation}
Generation quality is assessed at various computational budgets, measured in Number of Function Evaluations (NFE) $\in \{5,10,20,40,100\}$, across six complementary metrics: distributional distance (SWD~$\downarrow$~\cite{rabin2012sliced}, aFID~$\downarrow$~\cite{heusel2017fid}), sample quality and coverage (IS~$\uparrow$~\cite{salimans2016is}, Precision/Recall~$\uparrow$~\cite{kynkaanniemi2019prec}, Density/Coverage~$\uparrow$~\cite{naeem2020density}), and intra-set diversity (Div~$\uparrow$).

\subsection{Results}

\begin{figure}[ht!]
  \centering
  \begin{subfigure}[b]{0.45\linewidth}
    \includegraphics[width=\linewidth]{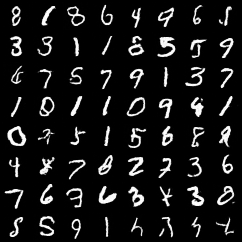}
    \caption{GPFN-stoch, NFE\,=\,20}
  \end{subfigure}\hfill
  \begin{subfigure}[b]{0.45\linewidth}
    \includegraphics[width=\linewidth]{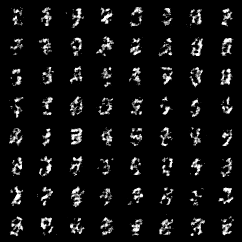}
    \caption{BFN-stoch, NFE\,=\,20}
  \end{subfigure}
  \caption{\textbf{Generated MNIST samples} (64 images per panel, NFE\,=\,20).}
  \label{fig:samples}
\end{figure}

\begin{figure}[ht]
  \centering
  \includegraphics[width=0.9\linewidth]{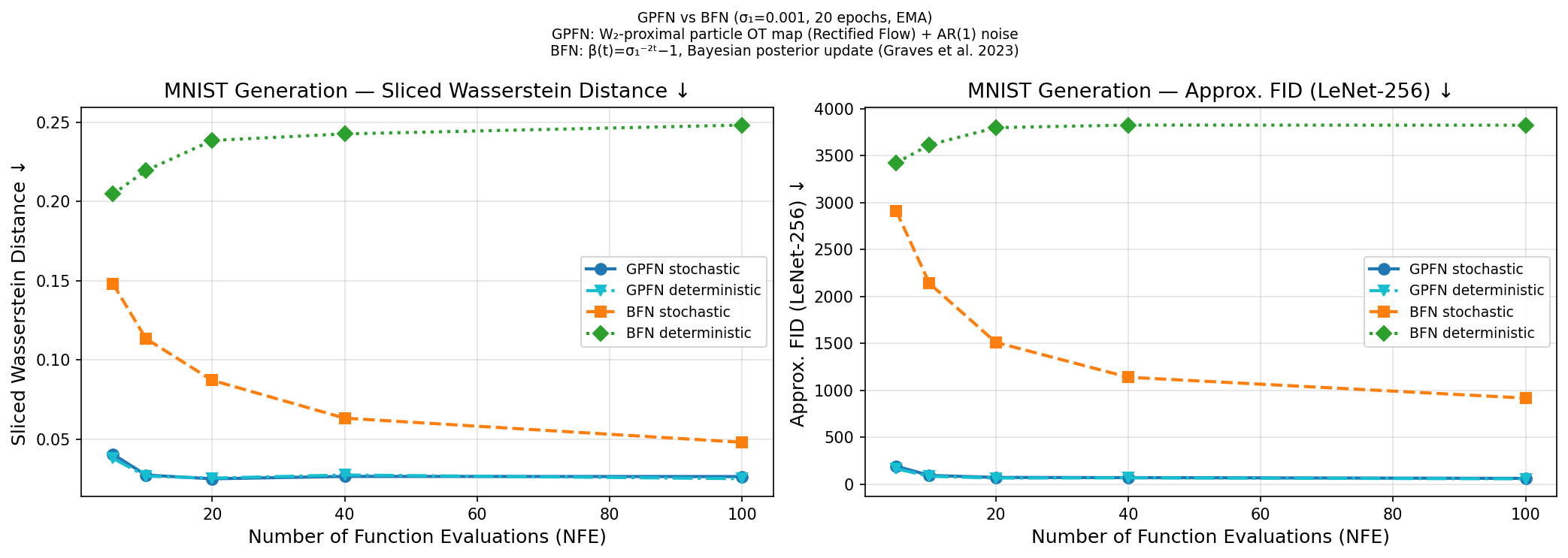}
  \caption{\textbf{Generation metrics vs.\ NFE budget.} GPFN achieves superior performance at a fraction of the steps required by BFN.}
  \label{fig:metric-curves}
\end{figure}

Table~\ref{tab:metrics} and Figures~\ref{fig:metric-curves}--\ref{fig:samples} summarise the results.

\begin{table}[ht]
\centering
\caption{\textbf{Generation metrics} for GPFN and BFN across different NFE budgets. Reported are Sliced Wasserstein Distance (SWD~$\downarrow$), approximate FID (aFID~$\downarrow$), Inception Score (IS~$\uparrow$), Precision (P~$\uparrow$), Recall (R~$\uparrow$), Density (D~$\uparrow$), Coverage (C~$\uparrow$), and Intra-set Diversity (Div~$\uparrow$). Best values per row for SWD and aFID are in \textbf{bold}.}
\label{tab:metrics}
\resizebox{\textwidth}{!}{%
\begin{tabular}{@{}lrrrrrrrrrrrr@{}}
\toprule
NFE & \multicolumn{3}{c}{GPFN-stoch} & \multicolumn{3}{c}{GPFN-det} & \multicolumn{3}{c}{BFN-stoch} & \multicolumn{3}{c}{BFN-det} \\
\cmidrule(lr){2-4} \cmidrule(lr){5-7} \cmidrule(lr){8-10} \cmidrule(lr){11-13}
    & SWD$\!\downarrow$ & aFID$\!\downarrow$ & IS/P/R/D/C/Div$\!\uparrow$ & SWD$\!\downarrow$ & aFID$\!\downarrow$ & IS/P/R/D/C/Div$\!\uparrow$ & SWD$\!\downarrow$ & aFID$\!\downarrow$ & IS/P/R/D/C/Div$\!\uparrow$ & SWD$\!\downarrow$ & aFID$\!\downarrow$ & IS/P/R/D/C/Div$\!\uparrow$ \\
    & ($\times10^{-2}$) & & & ($\times10^{-2}$) & & & ($\times10^{-2}$) & & & ($\times10^{-2}$) & & \\
\midrule
  5 & 4.06 & 193 & 7.5/0.91/0.83/0.73/0.78/0.41 & \textbf{3.80} & \textbf{166} & 7.7/0.87/0.86/0.71/0.78/0.42 & 14.82 & 2910 & 1.2/1.00/0.00/0.48/0.01/0.31 & 20.47 & 3423 & 1.0/1.00/0.00/0.40/0.00/0.00 \\
 10 & 2.75 &  95 & 8.3/0.86/0.90/0.73/0.84/0.40 & \textbf{2.68} &  \textbf{86} & 8.3/0.84/0.92/0.67/0.83/0.40 & 11.35 & 2139 & 1.8/1.00/0.00/0.76/0.02/0.32 & 21.95 & 3614 & 1.0/1.00/0.00/0.20/0.00/0.00 \\
 20 & \textbf{2.51} &  74 & 8.5/0.86/0.93/0.72/0.85/0.41 & 2.55 &  \textbf{67} & 8.6/0.84/0.95/0.65/0.84/0.40 &  8.72 & 1513 & 2.8/0.99/0.01/0.67/0.06/0.32 & 23.85 & 3798 & 1.0/1.00/0.00/0.20/0.00/0.00 \\
 40 & \textbf{2.66} &  72 & 8.5/0.87/0.94/0.75/0.88/0.40 & 2.75 &  \textbf{69} & 8.6/0.83/0.96/0.63/0.83/0.41 &  6.34 & 1141 & 3.7/0.93/0.14/0.51/0.10/0.33 & 24.26 & 3825 & 1.0/1.00/0.00/0.20/0.00/0.00 \\
100 & 2.65 &  64 & 8.6/0.86/0.95/0.74/0.89/0.41 & \textbf{2.52} &  \textbf{59} & 8.8/0.82/0.96/0.66/0.84/0.42 &  4.81 &  919 & 4.2/0.83/0.31/0.39/0.13/0.35 & 24.82 & 3824 & 1.0/1.00/0.00/0.20/0.00/0.00 \\
\bottomrule
\end{tabular}%
}
\end{table}

\subsection{Discussion}

\paragraph{GPFN achieves state-of-the-art performance at low NFE}
At NFE\,=\,20, the deterministic GPFN sampler (GPFN-det) achieves an aFID of 67, compared to 1513 for BFN-stoch. Even at an extremely low budget of NFE\,=\,5, GPFN-det reaches an aFID of 166, outperforming BFN-stoch at NFE\,=\,100 (aFID 919). This massive improvement stems from the $W_2$ proximal update inducing an optimal transport map on the particles, which coincides with the Euler integration of Rectified Flow~\cite{liu2022flow}. This allows GPFN to draw high-quality samples in very few steps.

\paragraph{Stabilized Stochastic Sampling}
The stochastic GPFN sampler (GPFN-stoch) uses an Ornstein-Uhlenbeck (AR(1)) process to update the noise. This provides principled stochasticity while avoiding the high variance of independent sampling, allowing it to closely match the deterministic sampler's performance across all budgets (aFID 64 at NFE\,=\,100).

\paragraph{BFN-det collapses}
As before, the deterministic BFN sampler fails completely (aFID $>$ 3400). Even though this custom sampler applies a $W_2$-proximal update, the underlying BFN probability paths lack the straight-line optimal transport geometry of GPFN. Without the stochastic injection defining the BFN forward process, the model is unable to transport probability mass correctly and cannot explore different modes of the data distribution. This is further evidenced by its Intra-set Diversity (Div) dropping to exactly 0.00 across all NFE budgets, indicating complete mode collapse to a single sample.

\paragraph{Sample Quality and Diversity}
The extended metrics (IS, Precision, Recall, Density, Coverage, and Diversity) further highlight the advantages of GPFN. GPFN maintains high Precision, Recall, Density, and Coverage across all NFE budgets, indicating that it generates high-quality samples that cover the data distribution well. For instance, at NFE\,=\,20, GPFN-det achieves a Precision of 0.84 and a Recall of 0.95, whereas BFN-stoch achieves a Precision of 0.99 but a Recall of only 0.01, indicating severe mode dropping. GPFN also maintains a healthy Intra-set Diversity ($\sim$0.40) across all budgets, confirming that the $W_2$ proximal update effectively transports the prior distribution to the full data distribution without losing modes.

\newpage

\section{Related Work}

\paragraph{Iterative refinement generative models}
Denoising Diffusion Probabilistic Models (DDPMs)~\cite{ho2020ddpm} and their improved variants~\cite{nicholdh2021improved} define the generative process as the reverse of a fixed Gaussian noise schedule, training a neural network to predict and remove noise at each step.
Score-based generative models~\cite{song2021score} unify this family under a continuous stochastic differential equation framework, where the score function is learned and used to drive a reverse-time SDE.
Flow Matching~\cite{lipman2022flow} and Rectified Flows~\cite{liu2022flow} replace stochastic dynamics with deterministic continuous normalizing flows, regressing on a simple linear interpolation field between noise and data.
Bayesian Flow Networks~\cite{graves2023bfn} take a different route: rather than transforming samples, they evolve a parametric belief distribution through a sequence of Bayesian posterior updates driven by noisy observations.
GPFNs generalize this last paradigm by replacing the fixed KL-divergence proximal step of BFNs with an arbitrary divergence. This subsumes BFNs as a special case while opening the framework to geometrically adaptive update rules.

\paragraph{Proximal algorithms and mirror descent}
Proximal point methods are a cornerstone of convex optimization~\cite{parikh2014proximal}: a sequence of iterates is produced by minimizing the objective augmented with a proximity term that penalizes deviation from the current point.
The choice of proximity measure determines the geometry of the updates; replacing the Euclidean penalty with a Bregman divergence yields mirror descent, which adapts to the geometry of the constraint set or the data distribution.
GPFNs instantiate this exact structure in the space of probability distributions, treating each belief update as a proximal step with a user-chosen divergence $D$.

\paragraph{Wasserstein gradient flows and the JKO scheme}
The Jordan--Kinderlehrer--Otto (JKO) scheme~\cite{jordan1998variational} interprets the Fokker--Planck equation as a Wasserstein gradient flow: each time step is a proximal point step in the space of probability distributions equipped with the $2$-Wasserstein metric $W_2$.
This variational perspective is directly mirrored in the $D=W_2^2$ instantiation of GPFNs: the closed-form Gaussian mean update derived in Appendix~\ref{app:impl} is the finite-step analogue of the continuous JKO iteration restricted to Gaussian families.
Crucially, the resulting particle update perfectly coincides with the Euler integration step of Rectified Flow~\cite{liu2022flow}. 
This connection suggests that GPFNs provide a principled foundation for Rectified Flows as discrete-time, parametric Wasserstein gradient flows, inheriting the favourable geometry of optimal transport for spatially structured data.

\newpage


\printbibliography

\appendix

\section{Implementation Details}\label{app:impl}

\subsection{GPFN Instantiation}

The belief family is restricted to axis-aligned Gaussians $p_t = \mathcal{N}(m_t, v_t I)$.
The proximal update~\eqref{eq:proximal-update} with $D = W_2^2$ and target $q = \delta_{\hat{x}_0}$ admits the closed-form update for both mean and variance:
\[
  m_{t+1} = \tau_t \hat{x}_0 + (1 - \tau_t) m_t, \qquad \sqrt{v_{t+1}} = (1 - \tau_t) \sqrt{v_t}
\]
where $\tau_t \in (0, 1)$ is the proximal step size.
This implies that the variance is strictly tied to the step size schedule via $v_t = \gamma_t^2$, where $\gamma_t = \prod_{i=0}^{t-1} (1 - \tau_i)$.
A cosine schedule $\gamma_t = \cos(\frac{t/T + s}{1 + s} \frac{\pi}{2}) / \cos(\frac{s}{1 + s} \frac{\pi}{2})$ with a small shift $s=0.008$ is chosen to prevent singularities near $t=0$, which yields $\tau_t = 1 - \gamma_{t+1}/\gamma_t$.
The network prediction $\hat{x}_0 = F_\theta(x_t,\, t/T)$ is obtained without self-conditioning to match the BFN baseline exactly, where $x_t \sim p_t$.
The training objective is pixel-level MSE computed on images linearly scaled to $[-1, 1]$ as defined in~\eqref{eq:training-loss}.

\paragraph{Sampling}
The $W_2$ proximal step on the belief distribution $p_t$ induces an optimal transport map on the particles $x_t \sim p_t$.
For the deterministic sampler (GPFN-det), this map is applied directly to the particles:
\[
  x_{t+1} = \tau_t \hat{x}_0 + (1 - \tau_t) x_t.
\]
Remarkably, this particle update perfectly coincides with the Euler integration step of Rectified Flow~\cite{liu2022flow}, revealing that GPFNs provide a principled derivation of Rectified Flow from the perspective of proximal belief updates.
For the stochastic sampler (GPFN-stoch), the belief mean $m_t$ is maintained and the noise $\varepsilon_t = (x_t - m_t)/\gamma_t$ is updated via an Ornstein-Uhlenbeck process $\varepsilon_{t+1} = \rho_t \varepsilon_t + \sqrt{1 - \rho_t^2} z$ with $\rho_t = \sqrt{1 - \tau_t}$ and $z \sim \mathcal{N}(0, I)$. This provides principled stochasticity while avoiding the high variance of independent sampling.

\subsection{BFN Baseline}

The BFN baseline architecture and training objective follow Graves et al.~\cite{graves2023bfn}. In addition to the original stochastic sampler, a custom BFN-det sampler is evaluated that applies the deterministic W2-proximal update directly to the BFN's predicted targets.
The Bayesian flow distribution is $\mu_t\!\mid\!x_0 \sim \mathcal{N}(\gamma x_0,\,\gamma(1\!-\!\gamma)I)$ with $\gamma(t)=1-\sigma_1^{2t}$ and $\sigma_1=0.001$.
The continuous-time training loss is $\mathcal{L}_\infty = -\ln(\sigma_1)\,\sigma_1^{-2t}\|x_0-\hat{x}_0\|^2$.

\subsection{Architecture}

Both models use an identical U-Net backbone with three encoder/decoder stages, base channel width 64, a bottleneck self-attention layer, and a sinusoidal time embedding.
The resulting parameter count is $4.032\text{M}$ for both models.
No self-conditioning is used in either model to ensure a perfectly fair comparison, meaning the architectures and parameter counts are literally identical.

\subsection{Training Hyperparameters}

All models are trained for 20 epochs with batch size 128.
The optimizer is AdamW~\cite{loshchilov2019adamw} with learning rate $2\!\times\!10^{-4}$ and cosine annealing.
Exponential moving average (EMA) of model weights is maintained with decay 0.999, and gradients are clipped at norm 1.0.

\subsection{Evaluation Protocol}

All feature-based metrics are computed by extracting 256-dimensional feature vectors from a lightweight LeNet classifier trained on MNIST.
Metrics are evaluated on 2{,}000 generated versus 2{,}000 real test images at each NFE value.
Intra-set diversity (Div) is computed as the mean $1-\text{SSIM}$ over 500 randomly sampled image pairs from the generated set.

\end{document}